\documentclass[10pt,twocolumn,letterpaper]{article}

\usepackage[pagenumbers]{cvpr} % To force page numbers, e.g. for an arXiv version

\usepackage{graphicx}
\usepackage{amsmath}
\usepackage{amssymb}
\usepackage{booktabs}
\usepackage{comment}
\usepackage{color}
\usepackage{mathtools}
\usepackage{scalerel}
\usepackage{algorithm}
\usepackage{algpseudocode}
\usepackage{epstopdf}
\usepackage{multirow}
\usepackage{array}
\newcolumntype{P}[1]{>{\centering\arraybackslash}p{#1}}
\newcolumntype{M}[1]{>{\centering\arraybackslash}m{#1}}
\usepackage[page]{appendix}
\usepackage{enumitem}

\usepackage[pagebackref,breaklinks,colorlinks]{hyperref}

\usepackage[capitalize]{cleveref}
\crefname{section}{Sec.}{Secs.}
\Crefname{section}{Section}{Sections}
\Crefname{table}{Table}{Tables}
\crefname{table}{Tab.}{Tabs.}

\def\cvprPaperID{*****} % *** Enter the CVPR Paper ID here
\def\confName{CVPR}
\def\confYear{2022}

\begin{document}

\title{Multi-label Classification with Partial Annotations using Class-aware Selective Loss}

\author{ Emanuel Ben-Baruch \hspace{0.1cm} Tal Ridnik \hspace{0.1cm}  \\ Itamar Friedman \hspace{0.1cm} Avi Ben-Cohen\hspace{0.1cm}  Nadav Zamir \hspace{0.1cm}  Asaf Noy \hspace{0.1cm}  Lihi Zelnik-Manor \vspace{0.1cm} \\ 
DAMO Academy, Alibaba Group\\
{\tt\small $\{$emanuel.benbaruch, tal.ridnik, itamar.friedman, avi.bencohen, nadav.zamir, asaf.noy, lihi.zelnik$\}$}\\  {\tt\small @alibaba-inc.com }
}

% \and
\maketitle

%%%%%%%%% ABSTRACT
\begin{abstract}
    \label{abstract}

Large-scale multi-label classification datasets are commonly, and perhaps inevitably, partially annotated. That is, only a small subset of labels are annotated per sample.
Different methods for handling the missing labels induce different properties on the model and impact its accuracy.
In this work, we analyze the partial labeling problem, then propose a solution based on two key ideas. 
First, un-annotated labels should be treated selectively according to two probability quantities: the class distribution in the overall dataset and the specific label likelihood for a given data sample.
We propose to estimate the class distribution using a dedicated temporary model, and we show its improved efficiency over a na\"ive estimation computed using the dataset's partial annotations.
Second, during the training of the target model, we emphasize the contribution of annotated labels over originally un-annotated labels by using a dedicated asymmetric loss.
% Experiments conducted on three partially labeled datasets, OpenImages, LVIS, and simulated-COCO, demonstrate the effectiveness of our approach. Specifically, with our novel selective approach, we achieve state-of-the-art results on OpenImages dataset. Code is available at 
% https://github.com/Alibaba-MIIL/PartialLabelingCSL.
With our novel approach, we achieve state-of-the-art results on OpenImages dataset (e.g. reaching 87.3 mAP on V6). In addition, experiments conducted on LVIS and simulated-COCO demonstrate the effectiveness of our approach.
Code is available at https://github.com/Alibaba-MIIL/PartialLabelingCSL.

\end{abstract}

%%%%%%%%% BODY TEXT
%------------------------------------------------------------------------
\section{Introduction}
\label{introduction} 

\begin{figure} [t!]
  \vspace{-0.1cm}
  \centering
  \includegraphics[scale=.42]{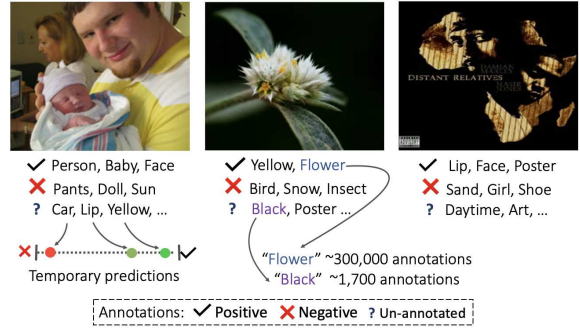}
  \vspace{-0.1cm}
  \caption{\textbf{Challenges in partial annotation.} 
%   Common classes may be insufficiently annotated.
  (1) ``Lip" and ``Yellow" are clearly present in the left image but were not annotated as positive labels. The middle and right images are annotated with ``Yellow" and ``Lip" respectively, while not being dominant labels in those images. (2) The deficiency of positive annotations is a key challenge: classes that frequently appear in images (e.g. ``Black", ``Lip") may be annotated much less  comparing with infrequent classes (``Flower", ``Guitar") (3) Most labels are un-annotated. How to exploit a temporary model's predictions for the un-annotated labels when training a target model?}
  \label{fig:lip_yellow}
  \vspace{-0.3cm}
\end{figure}

Recently, a remarkable progress has been made in multi-label classification \cite{gao2020multi, you2020cross, chen2019multi_MLGCN, lanchantin2020general}. Dedicated loss functions were proposed in \cite{ben2020asymmetric, tong2020distribution}, as well as transformers based approaches \cite{lanchantin2020general,liu2021query2label,cheng2021mltr}. 
% In spite of the recent achievements, handling partial annotations was not sufficiently investigated despite being a major challenge in multi-label classification problems. 
In many common cases, such as \cite{Kuznetsova_2020, gupta2019lvis, durand2019learning, DBLP:conf/nips/KunduT20, Huynh-mll:CVPR20}, as the amounts of samples and labels in the data increase, it becomes impractical to fully annotate each image. For example, OpenImages dataset \cite{Kuznetsova_2020} consists of 9 million training images and having 9,600 classes. An exhaustive annotation process would require annotating more than 86 billion labels. As a result, partially labeled data is inevitable in realistic large-scale multi-label classification tasks. A partially labeled image is annotated with a subset of positive labels and a subset of negative labels, where the rest un-annotated labels are considered as unknown (Figure \ref{fig:lip_yellow}). Typically, the majority of the labels are un-annotated. For example, on average, a picture in OpenImages is annotated with only 7 labels. Thus, the question of how to treat the numerous un-annotated labels may have a considerable impact on the learning process.

% \begin{figure} [t!]
%   \vspace{-0.1cm}
%   \centering
%   \includegraphics[scale=.16]{latex/Figures/lip_yellow_v6.png}
%   \vspace{-0.1cm}
%   \caption{\textbf{Challenges in partial annotation.} 
% %   Common classes may be insufficiently annotated.
%   (1) ``Lip" and ``Yellow" are clearly present in the left image but were not annotated as positive labels. The middle and right images are annotated with ``Yellow" and ``Lip" respectively, while not being dominant labels in those images. (2) The deficiency of positive annotations is a key challenge: classes that frequently appear in images (e.g. ``Black", ``Lip") may be annotated much less  comparing with infrequent classes (``Flower", ``Guitar") (3) Most labels are un-annotated. How to exploit a temporary model's predictions for the un-annotated labels when training a target model?}
%   \label{fig:lip_yellow}
%   \vspace{-0.3cm}
% \end{figure}

% \begin{figure*} [t!]
%   \centering
%   \includegraphics[scale=.3]{latex/Figures/Illustratiion.png}
%   \vspace{-0.3cm}
%   \caption{\textbf{Illustration of the challenges in the primary training modes.}  (a) \textit{Ignore} mode exploits only a subset of the samples which may lead to a limited decision boundary. (b) \textit{Negative} mode naively treats all un-annotated labels as negatives. It may produces suboptimal decision boundary as it adds noise of un-annotated positive labels. Also, annotated and un-annotated negative samples contribute similarly to the optimization. (c) Our approach aims at mitigating these drawbacks.}
%   \label{fig:illustration}
%   \vspace{-0.25cm}
% \end{figure*}

\begin{figure*} [t!]
  \centering
  \includegraphics[scale=.38]{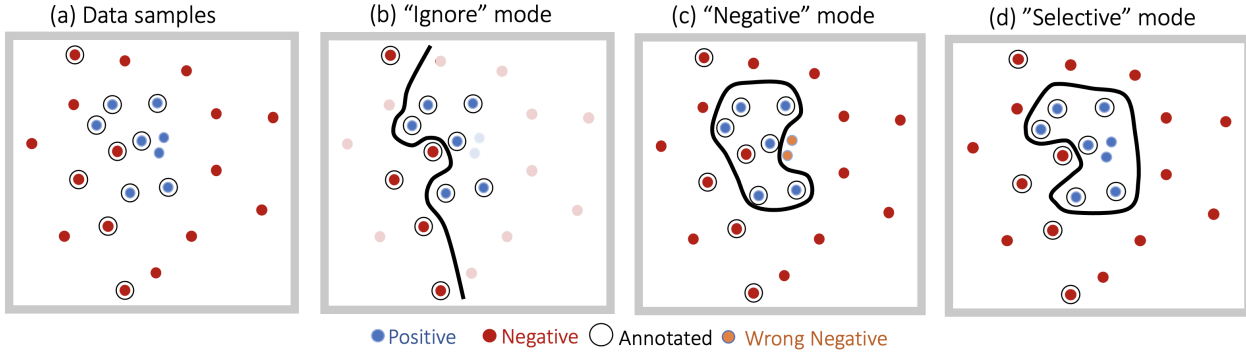}
  \vspace{-0.3cm}
  \caption{\textbf{Illustration of training modes for handling partial labeling.} (a) In a partially labeled dataset, only a portion of the samples are annotated for a given class. (b) \textit{Ignore} mode exploits only a subset of the samples which may lead to a limited decision boundary. (c) \textit{Negative} mode treats all un-annotated labels as negatives. It may produce suboptimal decision boundary as it adds noise of un-annotated positive labels. Also, annotated and un-annotated negative samples contribute similarly to the optimization process. (d) Our approach aims at mitigating these drawbacks by predicting the probability of a label being present in the image.}
  \label{fig:illustration}
  \vspace{-0.25cm}
\end{figure*}

The basic training mode for handling the un-annotated labels is simply to ignore their contribution in the loss function, as proposed in \cite{durand2019learning}. We denote this mode as \textit{Ignore}.
While ignoring the un-annotated labels is a reasonable choice, it may lead to a poor decision boundary as it exploits only a fraction of the data, see Figure \ref{fig:illustration}(b). Moreover, in a typical multi-label dataset, the probability of a label being negative is very high. Consequently, treating the un-annotated labels as negative may improve the discriminative power as it enables the exploitation of the entire data \cite{DBLP:conf/nips/KunduT20}. However, this training mode, denoted as \textit{Negative}, has two main drawbacks: adding label noise to the training, and inducing a high imbalance between negative and positive samples \cite{ben2020asymmetric}. This mode is illustrated in Figure \ref{fig:illustration}(c).

While treating the un-annotated labels as negative can be useful for many classes, it may significantly harm the learning of labels that tend to appear frequently in the images although not being sufficiently annotated. For example, color classes are labeled in only a small number of samples in OpenImages \cite{Kuznetsova_2020}, e.g. class ``Black" is annotated in 1,688 samples, which is only $0.02\%$ of the samples, while they are probably present in most of the images (see an example in Figure \ref{fig:lip_yellow}).
% In contrary, classes with probably low frequency such as "Boat" and "Snow" are annotated in more than 100,000 samples.
% Option 1: 
% For such classes, most of the loss gradients may be propagated from images that actually contain the label but were not being annotated with it.
% Option 2: 
Consequently, such classes are trained with many wrong negative samples.
Thus, it would be worthwhile to first identify the frequent classes in the data and treat them accordingly. While in fully annotated multi-label datasets (e.g. MS-COCO \cite{lin2014microsoft}) the class frequencies can be directly inferred by counting the number of their annotations, in partially annotated datasets it is not straightforward. Counting the number of positive annotations per class is misleading as the numbers are usually not proportional to the true class frequencies. In OpenImages, assumably infrequent classes like ``Boat" and ``Snow" are annotated in more than 100,000 samples, while frequent classes as colors are annotated in only $\sim$1,500 images.
Therefore, the class distribution is required to be estimated from the data. 

In this paper, we propose a \textit{Selective} approach that aims at mitigating the weaknesses raised by the primary training modes (Figure \ref{fig:illustration}). In particular, we will select one of the primary mode (\textit{Ignore} or \textit{Negative}) for each label individually by utilizing two probabilistic conditions, termed as \textit{label likelihood} and \textit{label prior}. The label likelihood quantifies the probability of a label being present in a specific image. The label prior represents the probability of a label being present in the data. To acquire a reliable label prior, we propose a method for estimating the class distribution. To that end, we train a classification model using \textit{Ignore} mode and evaluate it on a representative dataset. Then, when training the final model, to handle the high negative-positive imbalance, we adopt the asymmetric loss \cite{ben2020asymmetric}, which enables focusing on the hard samples, while at the same time controlling the impact from the positive and negative samples. We further suggest decoupling the focusing levels of the annotated and un-annotated terms in the loss to emphasize the contributions from the annotated negative samples. 

Extensive experiments were conducted on three datasets: OpenImages \cite{Kuznetsova_2020} (V3 and V6) and LVIS \cite{gupta2019lvis} which are partially annotated datasets with 9,600 and 1,203 classes, respectively. In addition, we simulated partially annotated versions of the MS-COCO \cite{lin2014microsoft} for exploring and verifying our approach. Results and comparisons demonstrate the effectiveness of our proposed scheme. Specifically, on OpenImages (V6) we achieve a state-of-the-art result of $87.34\%$ mAP score.
The contributions of the paper can be summarized as follows:
\begin{itemize}[leftmargin=0.4cm]
  \setlength{\itemsep}{0.2pt}
  \setlength{\parskip}{0.2pt}
  \setlength{\parsep}{0.2pt}
    \item Introducing a novel \textit{selective} scheme for handling partially labeled data, that treat each un-annotated label separately based on two probabilistic quantities: label likelihood and label prior. Our approach outperforms previous methods in several partially labeled benchmarks.
    % \item We identify the issue in calculating the class distribution in partial annotation and offer a simple yet effective approach for estimating the class distribution from the data.
    \item We identify a key challenge in partially labeled data, regarding the inaccuracy of calculating the class distribution using the annotations, and offer an effective approach for estimating the class distribution from the data.
    \item A partial asymmetric loss is proposed to dynamically control the impact of the annotated and un-annotated negative samples. 
\end{itemize}

\section{Related Work}
% \textbf{Partial labels.}
Several methods had been proposed to tackle the partial labeling challenge. \cite{durand2019learning} offered a partial binary cross-entropy (CE) loss to weigh each sample according to the proportion of known labels, where the un-annotated labels are simply ignored in the loss computation. 
In \cite{DBLP:conf/nips/KunduT20} they proposed to involve also the un-annotated labels in the loss, treating them as negative while smoothing their contribution by incorporating a temperature parameter in their sigmoid function. An interactive approach was presented in \cite{Huynh-mll:CVPR20} whose loss is composed of cross-entropy for the annotated labels and a smoothness term as a regularization. A curriculum learning strategy was also used in \cite{durand2019learning} to complete the missing labels.
Instead of using the same training mode for all classes, in this paper we propose adjusting the training mode, either as \textit{Ignore} or \textit{Negative} for each class individually, relying on probabilistic based conditions. Also, we introduce a key challenge in partial labeling, concerning the inability to infer the class distribution directly from the number of annotations, and suggest an estimation procedure to handle this. 

Other methods were proposed in \cite{yu2013largescale, yang2016improving, wu2018multilabel} to cope with partial labels, for example by a low-rank
empirical risk minimization \cite{yu2013largescale} or by learning structured semantic correlations \cite{yang2016improving}. However, they are not scalable to large datasets and their optimization procedures are not well adapted to deep neural networks. 

\begin{figure*} [t!]
  \centering
  \includegraphics[scale=.36]{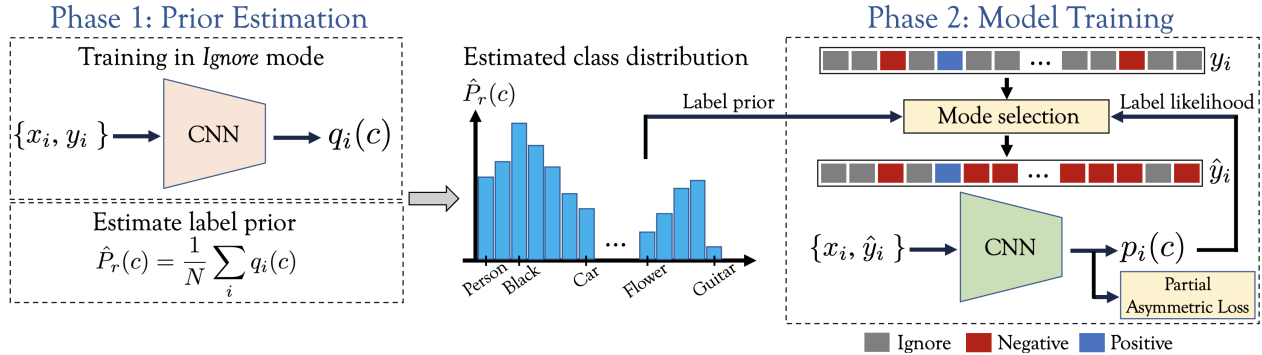}
  \vspace{-0.3cm}
  \caption{\textbf{Proposed approach.}  First, a class distribution estimation phase is performed to obtain a reliable label prior using a temporary network trained with the \textit{Ignore} mode. Then, the target model is trained using a \textit{Selective} approach which assigns a \textit{Negative} or \textit{Ignore} mode for each label based on its estimated prior and likelihood.}
  \label{fig:approach_scheme}
  \vspace{-0.0cm}
\end{figure*}

% Learning from noisy labels is a special case of partial labeling where the un-annotated labels are treated arbitrarily, for example by setting them as negative \cite{vahdat2017robustness, yan2019clusterfit, misra2016seeing}.
% Two close related fields are learning from noisy labels and positive-unlabeled (PU) learning. 
Positive Unlabeled (PU) is also related to partial labeling \cite{DBLP:journals/corr/abs-1811-04820,jiang2020improving,hammoudeh2020learning}. The difference is that PU learning approaches use only positive and un-annotated labels without any negative annotations. 

% The most relevant works to our paper are \cite{durand2019learning, DBLP:conf/nips/KunduT20, Huynh-mll:CVPR20}, which tackle the problem of partial labeling for large scale datasets using deep neural networks. However, in our approach we propose adjusting the training mode, either as \textit{Ignore} or \textit{Negative} for each label individually, relaying on probabilistic based conditions. Also, we introduce a key challenge in partial labeling, concerning the inability to infer the class distribution directly from the number of annotations, and suggest an estimation procedure to handle this. 

% Our proposed scheme relates to the fundamental objective function for training with partial annotations. It can be further integrated in an iterative approach as curriculum learning. However, it is not in the scope of this work.

%------------------------------s------------------------------------------
\section{Learning from Partial Annotations}
\label{section_2}

\subsection{Problem Formulation}
\label{section_2_modes}

Given a partially annotated multi-label dataset with $C$ classes, each sample $\mathbf{x} \in \mathcal{X}$, corresponding to a specific image, and is annotated by a label vector $\mathbf{y}=\{y_c\}_{c=1}^{C}$, where $y_c \in \{-1, 0, 1\}$ denotes whether the class $c$ is present in the image (`$1$'), absent (`$-1$') or unknown (`$0$'). 
For a given image, we denote the sets of positive and negative labels as $\mathcal{P}_{\mathbf{x}}=\{c|y_c=1\}$, and $\mathcal{N}_{\mathbf{x}}=\{c|y_c=-1\}$, respectively. The set of un-annotated labels is denoted by $\mathcal{U}_{\mathbf{x}}=\{c|y_c=0\}$. Note that typically, $| \mathcal{P}_{\mathbf{x}} \cup \mathcal{N}_{\mathbf{x}} | \ll |\mathcal{U}_{\mathbf{x}}|$.
A general form of the partially annotated multi-label classification loss can be defined as follows,
% \begin{equation}
%      \mathcal{L}(\mathbf{x})=
%      \smashoperator[r]{\sum_{\{c|y_c=1\}}}\mathcal{L}_{c}^{+}(\mathbf{x})
%      +\smashoperator[r]{\sum_{\{c|y_c=-1\}}}\mathcal{L}_{c}^{-}(\mathbf{x})
%      +\smashoperator[r]{\sum_{\{c|y_c=0\}}}\mathcal{L}_{c}^{u}(\mathbf{x})
% \end{equation}
\begin{equation}
     \mathcal{L}(\mathbf{x})=\sum_{c \in \mathcal{P}_{\mathbf{x}}}\mathcal{L}_{c}^{+}(\mathbf{x})
     +\sum_{c \in \mathcal{N}_{\mathbf{x}}}\mathcal{L}_{c}^{-}(\mathbf{x})
     +\sum_{c \in \mathcal{U}_{\mathbf{x}}}\mathcal{L}_{c}^{u}(\mathbf{x})
\end{equation}
where $\mathcal{L}_{c}^{+}(\mathbf{x})$, $\mathcal{L}_{c}^{-}(\mathbf{x})$ and $\mathcal{L}_{c}^{u}(\mathbf{x})$ are the loss terms of the positive, negative and un-annotated labels for sample $\mathbf{x}$, respectively.
Given a set of $N$ labeled samples $\{\mathbf{x}_i, \mathbf{y}_i\}_{i=1}^{N}$, our goal is to train a neural-network model $f(\mathbf{x}; \boldsymbol{\theta})$, parametrized by $\boldsymbol{\theta}$, to predict the presence or absence of each class given an input image. 
We denote by $\mathbf{p}=\{p_c\}_{c=1}^{C}$ the class prediction vector, computed by the model: $p_c=\sigma(z_c)$ where $\sigma(\cdot)$ is the sigmoid function, and $z_c$ is the output logit corresponding to class $c$.
% $\mathbf{p}=\sigma(f(\mathbf{x}; \boldsymbol{\theta}))$ where $\sigma(\cdot)$ is the sigmoid function.
% computed by $p_c=\sigma(z_c)$, where $\mathbf{z}=\{z_c\}_{c=1}^{C}$ is the vector of the output logits: $\mathbf{z}=f(\mathbf{x}; \boldsymbol{\theta})$ and $\sigma(z)$ is the sigmoid function. 

For example, applying the binary CE loss while considering only the annotated labels is defined by setting the loss terms as $\mathcal{L}_{c}^{+}(\mathbf{x}) = \log(p_c)$, $\mathcal{L}_{c}^{-}(\mathbf{x})=\log(1-p_c)$ and $\mathcal{L}_{c}^{u}(\mathbf{x})=0$.
% where $p_c$ is the network prediction for class $c$, computed by $p_c=\sigma(z_c)$.

\subsection{How to Treat the Un-annotated Labels?}
\label{sec:how_to_treat}
Typically, the number of un-annotated labels is much higher than the annotated ones. 
% For example, in OpenImages (V6) dataset, the average number of annotated labels per sample is 7, while the total number of classes is 9,600. 
Therefore,  the question of how to treat the un-annotated labels may have a considerable impact on the learning process. Herein, we will first define the two primary training modes and detail their strengths and limitations. Then, in light of these insights, we will propose a class aware mechanism which may better handle the un-annotated labels.

\textbf{Mode \textit{Ignore}}. The basic scheme for handling the un-annotated labels is simply to ignore them, as suggested in \cite{durand2019learning}. In this mode we set $\mathcal{L}_{c}^{u}(\mathbf{x})=0$.
%, \forall	c \in \mathcal{U}_{\mathbf{x}}$.
This way the training data is not contaminated with wrong annotations. However, its drawback is that it enables to use    only a subset of the data. For example, in OpenImages dataset, the number of samples with either positive or negative annotations for the class ``Cat" is only $\sim0.9\%$ of the training data. This may lead to a sub-optimal classification boundary when the annotated negative labels do not sufficiently cover the space of the negative class. See illustration in Figure \ref{fig:illustration}(b).

\textbf{Mode \textit{Negative}}. In typical multi-label datasets, the chance of a specific label to appear in an image is very low. For example, in the fully-annotated MS-COCO dataset \cite{lin2014microsoft}, a label is annotated as negative with a probability of $ \sim 0.96$. Based on this prior assumption, a reasonable choice would be to treat the un-annotated labels as negative, i.e. setting $\mathcal{L}_{c}^{u}(\mathbf{x})=\mathcal{L}_{c}^{-}(\mathbf{x})$.
%, \forall	c \in \mathcal{U}_{\mathbf{x}}$.
This working mode was also suggested in \cite{DBLP:conf/nips/KunduT20}. While this mode enables the utilization of the entire dataset, it suffers from two main limitations. First, it may wrongly annotate positive labels as negative annotations, adding label noise to the training. 
%Note that it may be especially harmful to frequent classes whose number of positive annotations is much lower than the actual number of images that contain the class. 
Secondly, this mode inherently triggers a high imbalance between negative and positive samples.
Balancing them, for example by down-weighting the contribution of the negative samples, may diminish the impact of the valuable annotated negative samples. These weaknesses are illustrated in Figure \ref{fig:illustration}(c).
% Moreover, for a given class, the annotated and un-annotated negative samples contribute similarly to the optimization. As will be argued in the following section, the impact of the annotated negative samples should be substantial, and not overwhelmed by the numerous un-annotated samples. These weaknesses are illustrated in \ref{fig:illustration}(b).
%  should be substantial, preventing them from being overwhelmed by the numerous un-annotated samples.

The question of which mode to choose has no unequivocal answer. It depends on various conditions and may have its origin in the annotation scheme used. In section \ref{sec:annotation_schemes}, we will show that different partial annotation procedures can lead to favor different loss modes (See Figure \ref{fig:impact_modes}).  
Moreover, as discussed in the next section, the used mode can influence each class differently, depends upon the class presence frequency in the data and the number of available annotations.
% It is further discussed in the next section. 
% Moreover, effectively, multi-label classification can be viewed as an extreme multi-task problem: each output neuron is independently activated by a sigmoid function which exhibits a binary classification task corresponding to a specific label. Consequently, the effective number of tasks can be enormous (e.g. $9,600$ labels in OpenImages or $1,203$ labels in LVIS.). Each label task is characterized by its unique nature and semantic level, and by different extent of imbalance between negative and positive samples. Therefore, the loss mode can affect each class differently. 

\subsection{Class Distribution in Partial Annotation}
% \subsection{Label Presence Frequency}
\label{sec:class_distribution}
As aforementioned, in multi-label datasets the majority of labels are present in only a small fraction of the data. For example, in MS-COCO, $89\%$ of the classes appear in less than $5\%$ of the data. 
% A similar trend can be assumed also in partially labeled datasets such as OpenImages and LVIS (though it cannot be quantitatively measured because the annotated ground-truth is partial). 
Thus, treating all un-annotated labels as negative may improve the discriminative power for many classes, as more real negative samples are involved in the training, while the added label noise is negligible. However, this may significantly harm the learning of classes whose number of positive annotations in the dataset is much lower than the actual number of samples they appear in. Consider the case of the class ``Person" in MS-COCO. It is present in $55\%$ of the data (45,200 samples). Now, suppose that only a subset of 1,000 positive annotations are available, and the rest are switched to negative. 
It means that during the training, most of the prediction errors are due to wrong annotations. In this case, the optimization will be degraded and the network confidence will be decayed considerably. Hence, it will be beneficial to first identify the frequent labels and handle them differently in the loss. 

\subsubsection{Positive Annotations Deficiency}
% \subsubsection{Why do we need to \textit{estimate} the class distribution?}
% \subsubsection{Class Distribution in Partial Annotation}
% \subsubsection{Using the Number of Positive Annotations is Insufficient}
To identify the frequent labels, we need to reliably acquire their distribution in the data.
% acquiring the class distribution in the data is needed. 
While in fully annotated datasets it can be easily obtained by counting the number of annotations per class and normalizing by the total number of samples, in partially annotated datasets it is not straightforward. While one may suggest counting the number of positive annotations for each class, the resulted numbers are misleading and are usually not proportional to the true class frequencies. For example, in OpenImages (V6), we found that many common and general classes which are frequently present in images are labeled with very few positive annotations. For example, general classes such as ``Daytime", ``Event" or ``Design" are labeled in only 1,709, 1,517 and 1,394 images (out of 9M), respectively. Color classes which massively appear in images are also rarely annotated. ``Black" and ``White" classes are labeled in only 1,688 and 1,497 images, respectively. We may assume that classes such as ``Daytime" or ``White" are present in much more than $0.02\%$ of the samples. Similarly, in LVIS dataset, the classes ``Person" and ``Shirt" are annotated in only 1,928 and 1,942 samples, respectively, while they practically appear in much more images (note that in MS-COCO, which shares the same images with LVIS, the class ``Person" appears in $55 \% $ of the samples).
% \begin{figure} [hbt!]
%   \vspace{-0.3cm}
%   \centering
%   \includegraphics[scale=.28]{latex/Figures/freq_classes.png}
%   \vspace{-0.3cm}
%   \caption{\textbf{Example of AP results for common classes.} Common classes are predicted better using \textit{Ignore} mode. For each class we indicate its number of positive annotations (in brackets). }
%   \label{fig:freq_classes}
%   \vspace{-0.7cm}
% \end{figure}

Note that the labels are not necessarily annotated according to their dominance in the image. In Figure \ref{fig:lip_yellow}, we show examples of three images and corresponding annotations of the classes ``Lip" and ``Yellow". As can be seen, the left image was not annotated with neither ``Lip" nor ``Yellow" although these labels are present and dominant in it. Also, ``Lip" is annotated in only 1,121 images which is highly deficient in view of the fact that the class ``Human face" is annotated in 327,899 images. 
% The class "Yellow" is annotated in only 1,324 images. 

According to the above-mentioned observations, the number of positive annotations cannot be used to measure the class frequencies in partially labeled datasets. In section \ref{sec: estimate_class_distribution}, we will propose a simple yet effective approach for estimating the class distribution from the data.

\section{Proposed Approach}
\label{section_3}

In this section we will present our method which aims at mitigating the issues raised in training partially annotated data. An overview of the proposed approach is summarized in Figure \ref{fig:approach_scheme}.
% The proposed approach is composed of two components: (1) partial asymmetric loss (P-ASL) decouples the un-annotated and annotated contributions, and (2) a selective mechanism that enable to assign a specific mode to each individual class according to two probabilistic expressions termed as label likelihood and label prior.
% \subsection{Asymmetric Loss for Partial Labeling}
% Title alternatives: Partial Assymetry Loss, Partial Assymetric Focusinig

To mitigate the high negative-positive imbalance problem, we adopt the asymmetric loss (ASL) proposed in \cite{ben2020asymmetric} as the base loss for the multi-label classification task. It enables to dynamically focus on the hard samples while at the same time controlling the contribution propagated from the positive and negative samples.
First, let us denote the basic term of the focal loss \cite{tsung2017focal} for a given class $c$, by:
\begin{equation}
    \mathcal{L}_{\scaleto{F}{3.5pt}}(p_c,\gamma) = (1-p_c)^{\gamma}\log p_c
\end{equation}
where $\gamma$ is the focusing parameter, which adjusts the decay rate of the easy samples. Then, we define the partially annotated loss as follows,
\begin{equation}
\label{eq:pasl}
\begin{split}
     \mathcal{L}(\mathbf{x}) & = \smashoperator[lr]{\sum_{c \in \mathcal{P}_{\mathbf{x}}}}\mathcal{L}_{\scaleto{F}{3.5pt}}(p_c, \gamma^{+})
     \\ &+\smashoperator[lr]{\sum_{c \in \mathcal{N}_{\mathbf{x}}}}\mathcal{L}_{\scaleto{F}{3.5pt}}(1-p_c, \gamma^{-})
     +\smashoperator[lr]{\sum_{c \in \mathcal{U}_{\mathbf{x}}}}\omega_c\mathcal{L}_{\scaleto{F}{3.5pt}}(1-p_c, \gamma^{u})
\end{split}
\end{equation}
where $\gamma^+$, $\gamma^-$ and $\gamma^u$ are the focusing parameters for the positive, negative and un-annotated labels, respectively. $\omega_c$ is the \textit{selectivity} parameter and it is introduced in section \ref{sec:selective_loss}.
We usually set $\gamma^{+} < \gamma^{-}$ to decay the positive term with a lower rate than the negative one because the positive samples are infrequent compared to the negative samples. In addition, as for a given class, the negative annotated samples are \textit{verified} ground-truth we are interested in preserving their contribution to the loss. Therefore, we suggest decoupling of the focusing parameter of the annotated negative labels from the un-annotated one, allowing us to set a lower decay rate for the annotated negative labels: $\gamma^{-} < \gamma^{u}$. This way, the impact of the annotated negative samples on establishing the classification boundary for each class is higher (see Figure \ref{fig:illustration}(d)). 
% Typical values can be set as $\gamma^+=0$, $\gamma^-=1$ and $\gamma^u=2$. 
We term this form of asymmetric loss as \textit{Partial}-ASL (P-ASL).

\subsection{Class-aware Selective Loss}\label{sec:selective_loss}
As described in section \ref{section_2_modes}, both \textit{Ignore} and \textit{Negative} modes are supported by inadequate assumptions for the partial annotation problem.
% While treating the un-annotated labels as negative enables utilizing the entire dataset, it complicates the training as it adds noise to the optimization.
In this section, we propose a selective approach for adjusting the mode per individual class. The core idea is to examine the probability of each un-annotated label being present in a given sample $\mathbf{x}$. Un-annotated labels that are suspected as positive will be ignored. The others will be treated as negative.

For that purpose, we define two probabilistic values: \textit{label likelihood} and \textit{label prior}, and detail their usage in the following section.
These two quantities are complementary to each other. The label likelihood enables to dynamically ignore the loss contribution of a label in a given image by inspecting its visual content. The label prior extracts useful information of the estimated class frequencies in the data and uses it regardless of the specific image content.

\begin{figure} [t!]
  \centering
  \includegraphics[scale=.35]{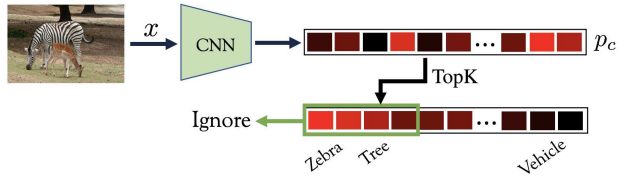}
  \vspace{-0.3cm}
  \caption{\textbf{Using the label likelihood.}  Un-annotated labels with highest network confidence may be related to positive items in the image. Thus, we ignore them in the loss computation.}
  \label{fig:likelihood_scheme}
  \vspace{-0.1cm}
\end{figure}

\textbf{Label likelihood.} 
Defined by the conditional probability of an un-annotated label $c$ of being positive given the image and the model parameters. i.e.
\begin{equation}
    P(y_c = 1 | \mathbf{x}; \boldsymbol{\theta}) ; \quad  \forall c \in \mathcal{U}_{\mathbf{x}}
\end{equation}
It can be simply estimated by the network prediction $\{p_c\}_{c \in \mathcal{U}_{\mathbf{x}}}$ throughout the training. A high $p_c$ may imply that the un-annotated label $c$ appears in the image, and treating it as negative may lead to an error. Accordingly, the label $c$ should be ignored. In practice, we allow for $K$ un-annotated labels with top prediction values to be ignored. i.e. 
\begin{equation}\label{eq:likelihood}
    \Omega_{L} = \big\{c \in \mathcal{U}_{\mathbf{x}} \mid	 c \in \text{TopK}(\{p_c\})\big\}
\end{equation}
where the $\text{TopK}(\cdot)$ operator returns the indices of the top $K$ elements of the input vector. The algorithm scheme is illustrated in Figure \ref{fig:likelihood_scheme}. Note that this implementation enables us to ``walk” on a continuous scale between the \textit{Negative} and \textit{Ignore} modes. Setting $K=0$ corresponds to \textit{Negative} mode, as no un-annotated label is ignored. $K=C$ equivalents to the \textit{Ignore} mode, as all un-annotated labels are ignored.

\textbf{Label prior.} Defined by the probability of a label $c$ being present in an image. It can also be viewed as the actual label presence frequency in the data. We are interested in the label prior for the un-annotated labels,
\begin{equation}
    P(y_c = 1) ; \quad  \forall c \in \mathcal{U}_{\mathbf{x}}.
     %{P}_r(c) = P(y_c = 1 |  \mathbf{x} \in \mathcal{X}).
\end{equation}
According to section \ref{sec:class_distribution}, the label prior should be estimated from the data, as the class distribution is hidden in partially annotated datasets. In the next section (\ref{sec: estimate_class_distribution}), we will introduce the scheme for estimating the label prior. Meanwhile, let us denote by $\hat{P}_r(c)$ the label prior estimator for class $c$. We are interested in disabling the loss contribution of labels with high prior values. These labels are formally defined by the following set,
\begin{equation}
    \Omega_{P} = \big\{c \in \mathcal{U}_{\mathbf{x}} \mid	  \hat{P}_r(c) > \eta \big\}
\end{equation}
where $\eta \in [0, 1]$ represents the minimum fraction of the data determining a label to be ignored.

Finally, we denote the set of labels whose loss contribution are ignored, as the union of the two previously computed sets,
\begin{equation}
    \Omega_{\text{Ignore}} = \Omega_{L} \cup	\Omega_{P}.
\end{equation}
Accordingly, we set the parameter $\omega_c$ in equation (\ref{eq:pasl}) as follows,
\begin{equation}\label{eq:omega_c}
\omega_c = 
    \begin{cases}
      0 \quad c \in \Omega_{\text{Ignore}}\\
      
      1  \quad c \notin \Omega_{\text{Ignore}}\\
    \end{cases}       
\end{equation}
% The algorithm is summarized in \ref{Algorithm}.
Note that we have explored other alternatives for implementing the label prior in the loss function. In particular, in appendix \ref{appendix:a} we compare a soft method that integrates the label prior by setting $\omega_c=\exp(-\alpha\hat{P}_r(c));   \forall c \notin \Omega_{L}$, and show that using a hard decision mechanism, as proposed in equation (\ref{eq:omega_c}), produces better results. 
% The method highlights is illustrated in Figure \ref{fig:approach_scheme}. In the following section we will detail the approach for estimating the label prior.

\subsection{Estimating the Class Distribution}
\label{sec: estimate_class_distribution}
We aim at estimating the class distribution in a representative dataset $\mathcal{X}$. For that, we first need to assess the presence of each class in every image in the data, i.e. we would like to first approximate the  probability of a class $c$ being present in an image $\mathbf{x} \in \mathcal{X}$:  $P(y_c= 1|\mathbf{x})$.
To that end, we propose training a model parametrized by $\boldsymbol{\theta}$, for predicting each class in a given image, i.e. $P(y_c= 1|\mathbf{x}; \boldsymbol{\theta})$.
Afterwards, the model is applied on the sample set $\mathcal{X}$ (e.g. the training data). The label prior can then be estimated by calculating the expectation, 
\begin{equation}
  \vspace{-0.15cm}
    P(y_c=1; \mathbf{\theta}) = \frac{1}{|\mathcal{X}|}\sum_{\mathbf{x} \in \mathcal{X}}P(y_c= 1|\mathbf{x}; \boldsymbol{\theta}).
  \vspace{-0.1cm}
\end{equation}

\begin{figure} [t!]
  \vspace{-0.1cm}
  \centering
  \includegraphics[scale=.38]{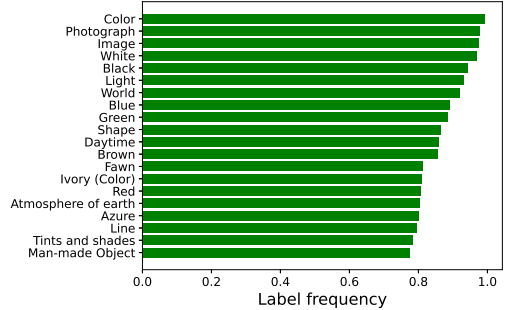}
  \vspace{-0.1cm}
  \caption{\textbf{Estimating the class distribution in OpenImages.} 
%   Common classes may be insufficiently annotated.
  Top 20 frequent classes estimated by the \textit{Ignore} model. Top classes are all related to common labels such as colors or general concepts.}
  \label{fig:estimate_classs_opim}
  \vspace{-0.3cm}
\end{figure}

For estimating the label priors, we train the model in \textit{Ignore} mode. While the discriminative power of the \textit{Negative} mode may be stronger for majority of the labels, it may fail to provide a reliable prediction values for frequent classes with small number of positive annotations. Propagating abundance of gradient errors from wrong negative annotations will decay the expected returned prediction for those classes and will fail to approximate $P(y_c= 1|\mathbf{x})$.
Consequently, our suggested estimation for the class distribution is given by,
\begin{equation} \label{eq:prior}
    \hat{P}_r(c) = P(y_c=1; \mathbf{\boldsymbol{\theta}_{\text{Ignore}}}),
\end{equation}
where $\boldsymbol{\theta}_{\text{Ignore}}$ denotes the model parameters trained in \textit{Ignore} mode.
In section \ref{sec:results_estimate_class_prior}, we will empirically show the effectiveness of the \textit{Ignore} mode in ranking the class frequencies and the inapplicability of the \textit{Negative} mode to do that.
% i.e we will show that $P(y_c= 1; \boldsymbol{\theta}_{\text{Ignore}}) \approx P(y_c= 1)$, while $P(y_c= 1; \boldsymbol{\theta}_{\text{Negative}}) \not\approx P(y_c= 1)$, where $\boldsymbol{\theta}_{\text{Negative}}$ denotes the model parameters trained in \textit{Negative} mode.
To qualitatively show the estimation effectiveness, we present in Figure \ref{fig:estimate_classs_opim} the top 20 frequent classes in OpenImages (V6) as estimated by our proposed procedure. Note that all the top classes are commonly present in images such as colors (``White", ``Black", ``Blue" etc.) or general classes such as ``Photograph", ``Light", ``Daytime" or ``Line". In appendix \ref{appendix:c}, we show the next top 60 estimated classes. Also, in appendix \ref{appendix:d}, we provides the top 20 estimated frequent classes for LVIS dataset.

\section{Experimental Study  }
\label{experiments}
In this section, we will experimentally demonstrate the insights discussed in the previous sections. We will mainly utilize the fully annotated MS-COCO dataset \cite{lin2014microsoft} to validate and demonstrate the effectiveness of our approach by simulating partial annotation under specific case studies. The evaluation metric used in the experiments is the mean average precision (mAP). Training details are provided in appendix \ref{appendix:training_details}.
% In this section, we will experimentally demonstrate the main insights discussed in the previous sections as well as presenting results and ablation study of the proposed approach. We will show results on three publically available datasets: OpenImages \cite{Kuznetsova_2020}, and LVIS \cite{gupta2019lvis} are partially annotated multi-label datasets. In addition, we utilize the fully annotated MS-COCO dataset \cite{lin2014microsoft} to validate and demonstrate the effectiveness of our approach by simulating partial annotation under specific case studies. The evaluation metric used in the experiments is the mean average precision (mAP). In particular, we report the standard per-class mAP denoted as mAP(C), and overall mAP denoted as mAP(O), which considers the number of samples in each class. The training details and the loss hyper-parameters used are provided in appendix \ref{appendix:training_details}.

\subsection{Impact of Annotation Schemes}
\label{sec:annotation_schemes}
% Comparison between two-three annotation schemes using simulated COCO dataset.
As aforementioned in section \ref{sec:how_to_treat}, the scheme used for annotating the dataset can substantially induce the learning process. Specifically, the choice of how to treat the un-annotated labels is highly influenced by the annotation scheme. To demonstrate that, we simulate two partial annotation schemes on the original fully annotated MS-COCO dataset \cite{lin2014microsoft}. 
MS-COCO includes 80 classes, 82,081 training samples, and 40,137 validation samples, following the 2014 split. The two simulated annotation schemes are detailed as follows:

% \textbf{MS-COCO Dataset.} A widely used benchmark for evaluating multi-label classification tasks. It includes 80 classes, and we use the 2014 split which contains 82,081 images for training and 40,137 images for validation. The dataset is fully annotated.

\noindent\textbf{Fixed per class (FPC).} 
% we keep the same number of positive and negative annotations for every class. In particular, 
For each class, we randomly sample a fixed number of positive annotations, denoted by $N_s$, and the same number of negative annotations. The rest of the annotations are dropped. 
% $N_s$ is predefined. In the case of classes with less than $N_s$ annotations, we simply keep all annotations. 
% For example, FPC-1000/1000 refe    rs to the case where we randomly sample for each class a maximum of $1,000$ positive and negative annotations. 

\noindent\textbf{Random per annotation (RPA).}
We omit each annotation with probability $p$. Note that this simulation preserves the true class distribution of the data. 

\begin{figure}[t!]
\begin{subfigure}[a]{.23\textwidth}
  \centering
  \includegraphics[width=1.0\linewidth]{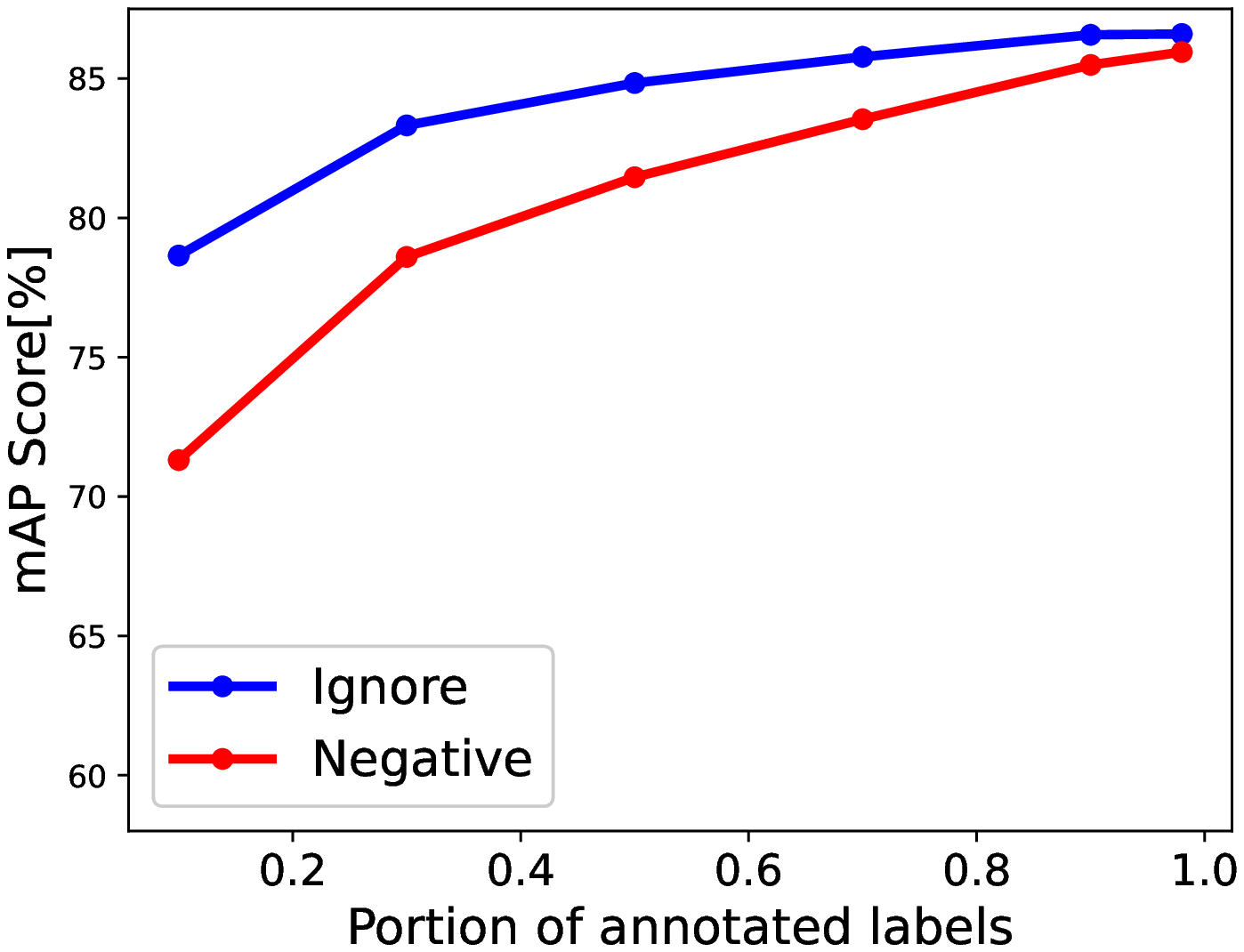}
  \caption{}
\end{subfigure}%
\begin{subfigure}[h]{.23\textwidth }
  \centering
  \includegraphics[width=1.0\linewidth]{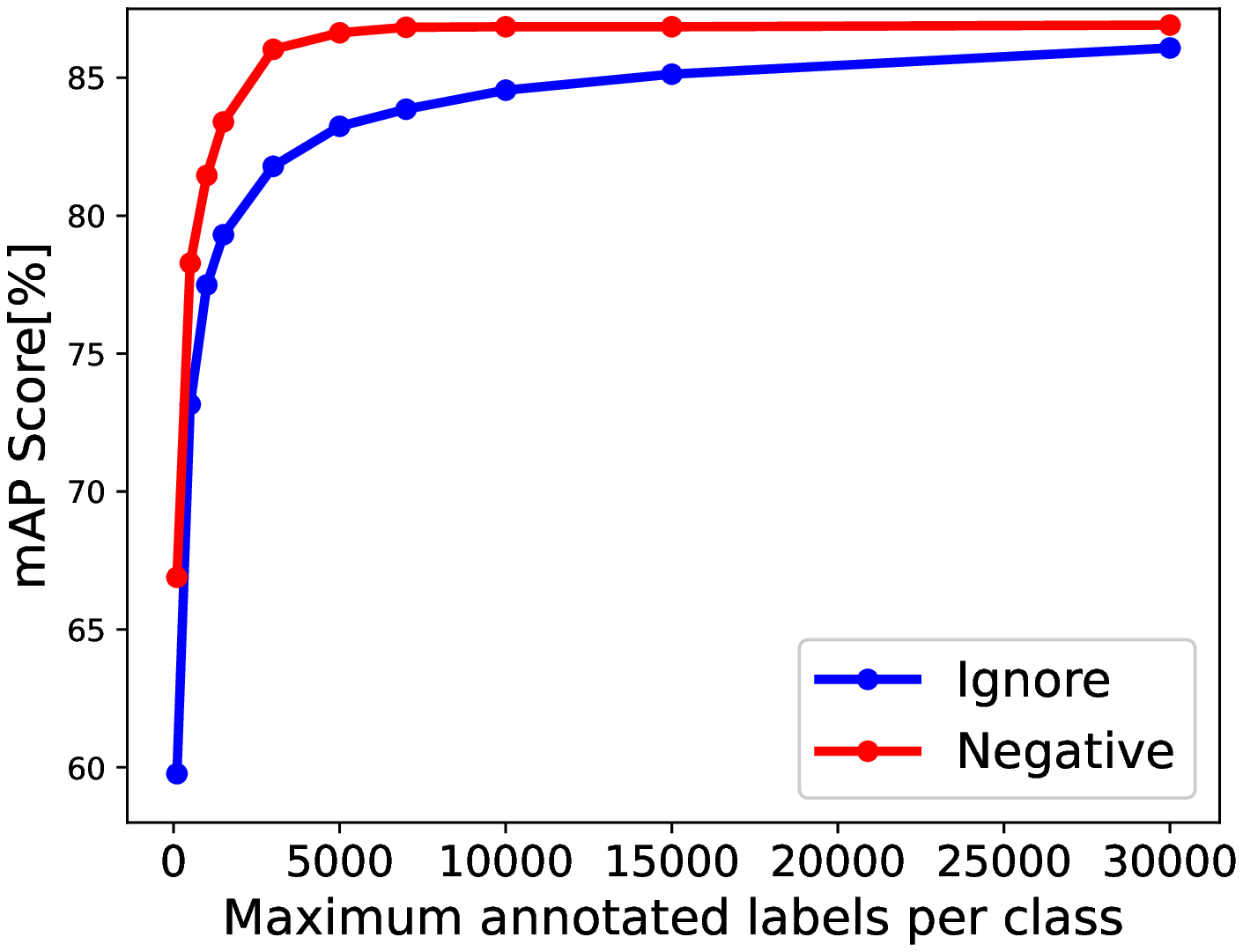}
  \caption{}
\end{subfigure}
\caption{\textbf{Impact of annotation schemes.} mAP results obtained using the RPA (a) and the FPC (b) simulation schemes for each primary mode. While in RPA, \textit{Ignore} mode consistently shows better results, in FPC, the \textit{Negative} mode is superior.}
\label{fig:impact_modes}
\vspace{-3mm}
\end{figure}

In Figure \ref{fig:impact_modes}, we show results obtained using each one of the simulation schemes for each primary mode (\textit{Ignore} and \textit{Negative}) while varying $N_s$ and $p$ values. As can be seen, while in RPA (Figure \ref{fig:impact_modes}(a)), the \textit{Ignore} mode consistently shows better results, in FPC (Figure \ref{fig:impact_modes}(b)), the \textit{Negative} mode is superior. 
% Note that as we keep more annotated labels (by either increasing $N_s$ or decreasing $p$), the gap between the two training modes is reduced, catching up to the maximal result.
Note that as we keep more of the annotated labels (by either increasing $N_s$ or decreasing $p$), the gap between the two training modes is reduced, catching the maximal result.
The phenomenons observed in the two case studies we simulated are also related in real practical procedures for partially annotating multi-label datasets. While in the FPC simulation, the class distribution is completely vanished and cannot be inferred by the number of positive annotations ($N_s$ for  $c=1,..., C$), the RPA scheme preserves the class distribution.

\begin{figure} [t!]
  \vspace{-0.2cm}
  \centering
  \includegraphics[scale=.38]{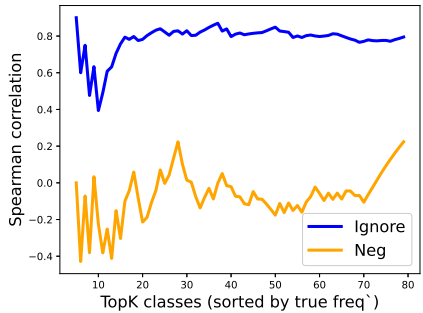}
  \vspace{-0.1cm}
%   \caption{\textbf{Spearman correlation between the true class distribution and the estimated distribution, computed on MS-COCO for the FPC simulation scheme.} 
  \caption{\textbf{Spearman correlation between the true class distribution and the estimated distribution.} 
  Unlike the \textit{Negative} mode, training a model using \textit{Ignore} mode is well suited for estimating the class distribution.}
  \label{fig:coco_spearman}
  \vspace{-0.1cm}
\end{figure}

\subsection{Estimating the Label Prior}
\label{sec:results_estimate_class_prior}
To demonstrate the estimation quality of the class distribution obtained by the approach proposed in section \ref{sec: estimate_class_distribution}, we follow the FPC simulation scheme applied on the MS-COCO dataset (as described in section \ref{sec:annotation_schemes}), where a constant number of 1,000 annotations remained for each class. Because MS-COCO is a fully annotated dataset, we can compare the estimated class distribution (i.e. the label prior) to the true class distribution inferred by the original number of annotations. In particular, we measure the similarity between the original class frequencies and the estimated ones using the Spearman correlation test. In figure \ref{fig:coco_spearman}, we show the Spearman correlation scores while varying the number of top-ranked classes. We also show the results obtained with \textit{Negative} mode as a reference. Specifically, the Spearman correlation computed over all the 80 classes, with the estimator obtained using the \textit{Ignore} mode is $0.81$, demonstrating the estimator's effectiveness. In the next section, we will show how it benefits the overall classification results. Also, in appendix \ref{appendix:b} we present the top frequent classes measured by our estimator and compare them to those obtained by the original class frequencies in MS-COCO.

\section{Benchmark Results}
\label{benchmark results}

In this section, we will report our main results on the partially annotated multi-label datasets: OpenImages \cite{Kuznetsova_2020}, and LVIS \cite{gupta2019lvis}. The results on MS-COCO dataset are presented in appendix \ref{appendix:b}. We will present a comparison to previous methods which handle partial annotations, among other baseline approaches in multi-label classification. The evaluation metric used in the experiments is the mean average precision (mAP). In particular, we report the standard per-class mAP denoted as mAP(C), and overall mAP denoted as mAP(O), which considers the number of samples in each class. The training details and the loss hyper-parameters used are provided in appendix \ref{appendix:training_details}.
\begin{table}[t!]
\centering
\begin{tabular}{ |p{3.6cm}||P{1.3cm}|P{1.3cm}|}
 \hline
 Method & mAP(C) &  mAP(O)\\
 \hline
%  CE, \textit{Ignore}                                      & 86.12    & 93.41 \\
 CE, \textit{Ignore}                                      & 85.38    & 93.15 \\
 wCE \cite{durand2019learning}, \textit{Ignore}           & 85.22    & 93.05 \\
%  ASL \cite{ben2020asymmetric}, \textit{Ignore}            & 86.22    & 93.45 \\
 \hline
 CE, \textit{Negative}                                    & 85.35    & 91.14 \\
 SE(LS) \cite{DBLP:conf/nips/KunduT20}, \textit{Negative}     & 85.70    & 91.20 \\
 ASL \cite{ben2020asymmetric}, \textit{Negative}          & 85.85    & 91.29 \\
%  P-ASL, \textit{Negative}                                 & 86.28    & 92.34 \\
 \hline
 P-ASL, \textit{Negative}       & 86.28    & 92.34 \\
 P-ASL, \textit{Selective} ($\Omega_L$)       & 86.36    & 93.25 \\
 P-ASL, \textit{Selective} ($\Omega_P$)       & 86.46    & 93.27 \\
 P-ASL, \textit{Selective} ($\Omega_{\text{Ignore}}$)       & \textbf{86.72}    & \textbf{93.57} \\
%  P-ASL, \textit{Selective}       & \textbf{86.72}    & \textbf{93.57} \\
 \hline
\end{tabular}
\caption{\textbf{OpenImages (V6) results.} The \textit{Selective} approach with P-ASL improves both mAP(C) and mAP(O) scores.}
\label{table: opim_v6}
\vspace{-0.2cm}
\end{table}

% \vspace{0.8cm}
\begin{table}[t!]
\centering
\begin{tabular}{ |p{3.6cm}||P{1.3cm}|P{1.3cm}|}
 \hline
 Backbone & mAP(C) &  mAP(O)\\
 \hline
 OFA-595  \cite{cai2020once}      & 85.40 & 92.87 \\
 ResNet-50 \cite{he2016deep}             & 86.15 & 93.16 \\
 TResNet-M \cite{ridnik2021tresnet}      & 86.72    & 93.57 \\
 TResNet-L \cite{ridnik2021tresnet}      & \textbf{87.34}    & \textbf{93.77} \\
 \hline
\end{tabular}
\caption{\textbf{OpenImages (V6) results for different backbones.} Using TResNet-L model we achieve top results on OpenImages V6.}
\label{table: opim_v6_backbones}
\vspace{-0.3cm}
\end{table}

% \vspace{0.8cm}
\begin{table*}[t!]
\centering
\begin{tabular}{ |p{3.6cm}||M{1.3cm}|M{1.3cm}|M{1.3cm}| M{1.3cm}| M{1.3cm}| M{1.3cm}|}
 \hline
 Method & Group 1 &  Group 2 &  Group 3 &  Group 4 &  Group 5 &  All classes \\
 \hline
 Latent Noise (visual) \cite{misra2016seeing}      & 69.37 & 70.41 & 74.79 & 79.20 & 85.51 & 75.86 \\
 CNN-RNN \cite{wang2016cnnrnn}                     & 68.76 & 69.70 & 74.18 & 78.52 & 84.61 & 75.16 \\
 Curriculum Labeling \cite{durand2019learning}     & 70.37 & 71.32 & 76.23 & 80.54 & 86.81 & 77.05 \\
 IMCL \cite{Huynh-mll:CVPR20}                      & 70.95 & 72.59 & 77.64 & 81.83 & 87.34 & 78.07 \\
 \hline
%  P-ASL, \textit{Selective} (ours)  & \textbf{75.34} & \textbf{80.49} & \textbf{86.66} & \textbf{88.92} & \textbf{91.53} & \textbf{84.58} \\
%  P-ASL, \textit{Selective} (ours)  & \textbf{73.74} & \textbf{78.90} & \textbf{85.27} & \textbf{87.85} & \textbf{90.72} & \textbf{83.31} \\
 P-ASL, \textit{Selective} (ours)  & \textbf{73.19} & \textbf{78.61} & \textbf{85.11} & \textbf{87.70} & \textbf{90.61} & \textbf{83.03} \\
 \hline
\end{tabular}
\caption{\textbf{Results for OpenImages (V3).} Comparing the mAP score obtained using our \textit{Selective} approach to previous multi-label classification methods.}
\label{table: opim_v3}
\vspace{-0.7cm}
\end{table*}

\vspace{-0.1cm}
\subsection{OpenImages V6}
\vspace{-0.1cm}
% \textbf{OpenImages V6.} 
Openimages V6 is a large-scale multi-label dataset \cite{Kuznetsova_2020}, consists of 9 million training images, 41,620 validation samples, and 125,456 test samples. It is a partially annotated dataset, with 9,600 trainable classes.
% In the following experiments, unless stated otherwise, the hyper-parameters were set as follows: $\eta=0.05$, $K=200$, $\gamma_u=7$, $\gamma_-=2$ and $\gamma_+=1$.
In Table \ref{table: opim_v6}, we present the mAP results obtained by our proposed \textit{Selective} method and compare them to other approaches. Interestingly, \textit{Ignore} mode produces better results than \textit{Negative} mode, as OpenImages contains many under-annotated frequent classes such as colors and other general classes (see Figure \ref{fig:estimate_classs_opim}). Using \textit{Negative} mode adds a massive label noise and harms the learning of many common classes. In Table \ref{table: opim_v6_backbones}, we present results for different network architectures. Specifically, using TResNet-L \cite{ridnik2021tresnet}, we achieve state-of-the-art result of $87.34$ mAP score.
  
\begin{figure} [t!]
  \vspace{-0cm}
  \centering
  \includegraphics[scale=.36]{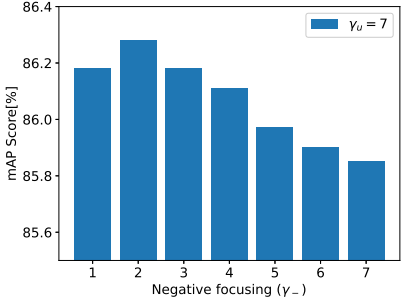}
  \vspace{-0.1cm}
  \caption{\textbf{Impact of decoupling the focusing parameters.} 
  We set the un-annotated focusing to $\gamma_u=7$ and varied the annotated negative focusing $\gamma_-$.}
  \label{fig:pasl_opim}
  \vspace{-0.5cm}
\end{figure}

%  \vspace{-0.1cm}
% \textbf{Ablation Study.}
To show the impact of decoupling the focusing parameters of the annotated and un-annotated loss terms in P-ASL as proposed in equation (\ref{eq:pasl}), we varied the negative focusing parameter $\gamma_-$, while fixing $\gamma_u=7$. The results are presented in Figure \ref{fig:pasl_opim}. The case of $\gamma_-=7$ represents the standard ASL \cite{ben2020asymmetric}. As can be seen, the mAP score increases as we lower $\gamma_-$, up to 2. It indicates that lowering the decay rate for the annotated negative term boosts their contribution to the loss.

\begin{figure} [t!]
  \vspace{-0.3cm}
  \centering
  \includegraphics[scale=.36]{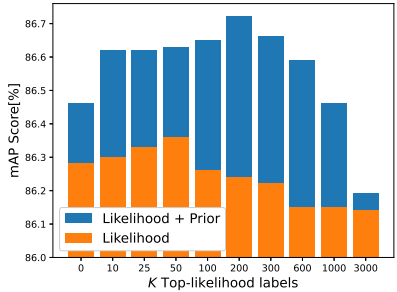}
  \vspace{-0.1cm}
  \caption{\textbf{Ablation study of the \textit{Selective} approach components.} 
  mAP results are shown for different numbers of top likelihood labels, $K$. We show results for the case of using only the likelihood condition $\Omega_L$, and with both conditions $\Omega_L \cup \Omega_P$.}
  \label{fig:selective_ablation_opim}
  \vspace{-0.2cm}
\end{figure}

In Figure \ref{fig:selective_ablation_opim}, we show the mAP scores while varying the number of top likelihood classes, $K$ as defined in equation (\ref{eq:likelihood}). Note that setting $K=0$ is equivalent to use $\textit{Negative}$ mode.  Training with high enough $K$ becomes similar to training using $\textit{Ignore}$ mode.
% (for $K$=3,000 we achieve mAP=$86.15$).
The highest mAP results are obtained with both the likelihood and prior conditions.

\subsection{OpenImages V3}
 \vspace{-0.1cm}
% \textbf{OpenImages V3.}
To be compatible with previously published results, we used the OpenImages V3 which contains 5,000 trainable classes. We follow the comparison setting described in \cite{Huynh-mll:CVPR20}. Also, for a fair comparison we used the ResNet-101 \cite{he2016deep} backbone, pre-trained on the ImageNet dataset. 
% The hyper-parameters were set as follows: $\eta=0.05$, $K=200$, $\gamma_u=4$, $\gamma_-=2$ and $\gamma_+=1$.
In Table \ref{table: opim_v3}, we show the mAP score results obtained using previous approaches and compared them to our $\textit{Selective}$ method. As shown, our method significantly outperforms previous approaches that deal with partial annotation in a multi-label setting.

 \vspace{-0.1cm}
\subsection{LVIS}
 \vspace{-0.1cm}
LVIS is a partially labeled dataset originally annotated for object detection and image segmentation, that was adopted as a multi-label classification benchmark. It consists of 100,170 images for training and 19,822 images for testing. It contains 1,203 classes. In Table \ref{table:lvis}, we present a comparison between different approaches on the LVIS dataset. As can be seen, in this case, the \textit{Negative} mode is better, compared to the \textit{Ignore} mode. This can be related to the fact that most of the labels are related to specific objects which do no appear frequently in the images. The most frequent class is ``Person". Therefore we also added its average precision to Table \ref{table:lvis}. Note that the \textit{Ignore} model better learns the class ``Person" compared to the one trained with \textit{Negative} mode. Using the P-ASL with the \textit{Selective} mode, we were able to obtain superior mAP results as well as top average precision even for the most frequent class, ``Person". 

% \vspace{0.8cm}
\begin{table}[t!]
\centering
\begin{tabular}{ |p{2.8cm}||M{1.0cm}|M{1.0cm}| M{1.0cm}|}
 \hline
 Method & mAP(C) &  mAP(O) & Person-AP\\
 \hline
 CE, \textit{Ignore}                                      & 74.49    & 95.70 & 99.81 \\
 wCE \cite{durand2019learning}, \textit{Ignore}           & 74.15    & 95.20 & 99.80\\
%  ASL \cite{ben2020asymmetric}, \te    xtit{Ignore}            & 74.63    & 95.81 & -\\
 \hline
 CE, \textit{Negative}                                    & 77.82    & 96.66 & 97.20 \\
 SE \cite{DBLP:conf/nips/KunduT20}, \textit{Negative}     & 77.81    & 96.60 & 97.28 \\
 ASL \cite{ben2020asymmetric}, \textit{Negative}          & 78.32    & 96.77 & 98.43\\
%  P-ASL, \textit{Negative}                                 & 86.28    & 92.34 \\
 \hline
 P-ASL,\textit{Selective}                              & \textbf{78.57}    & \textbf{96.80} & \textbf{99.81}\\
 \hline
\end{tabular}
\caption{\textbf{Results for LVIS.} The \textit{Selective} approach with P-ASL improves both mAP(C) and mAP(O) scores. Also, it provides top result for the frequent class "Person".}
\label{table:lvis}
\end{table}

% \subsubsection{MS-COCO}
% Results on MS-COCO dataset are presented in appendix \ref{appendix:b}.

%------------------------------------------------------------------------
\section{Conclusion}
\label{conclusion}
In this paper, we presented a novel technique for handling partially labeled data in multi-label classification. We observed that ignoring the un-annotated labels in the loss or treating them as negative should be determined individually for each class. We proposed a selective mechanism that uses the label likelihood computed throughout the training, and the label prior which is obtained by estimating the class distribution from the data. The un-annotated labels are further softened via a partial asymmetric loss. Extensive experiments analysis shows that our proposed approach outperforms other previous methods on partially labeled datasets, including OpenImages, LVIS, and simulated-COCO. 

%------------------------------------------------------------------------
% \section{Draft notes}
% \input{latex/draft-notes}

%%%%%%%%% REFERENCES
{\small
\bibliographystyle{ieee_fullname.bst}
\bibliography{egbib.bib}
}

\clearpage
\appendix
\begin{appendices}

\section{Training Details}\label{appendix:training_details}
Unless stated otherwise, all experiments were conducted with the following training configuration. As a default, we used  the TResNet-M model \cite{ridnik2021tresnet}, pre-trained on ImageNet-21k dataset \cite{ridnik2021imagenet21k}. The model was fine-tuned using Adam optimizer \cite{kingma2017adam} and 1-cycle cosine annealing policy
\cite{smith2018disciplined} with a maximal learning rate of 2e-4 for training OpenImages and MS-COCO, and 6e-4 for training LVIS. 
% We used Cutout \cite{Cutout} with a probability of 0.5,
We used true-weight-decay \cite{loshchilov2017decoupled} of 3e-4 and
standard ImageNet augmentations. For fair comparison to previously published results on OpenImages V3, we also trained a ResNet-101 model, pre-trained on ImageNet.

In the OpenImages experiments we used the following hyper-parameters: $\eta=0.05$, $K=200$, $\gamma_u=7$, $\gamma_-=2$ and $\gamma_+=1$. In LVIS we used: $\gamma_u=1$, $\gamma_-=0$ and $\gamma_+=0$.

\section{Soft Label Prior}\label{appendix:a}
Herein, we will explore a soft alternative for integrating the label prior in the loss. 
We follow equation (\ref{eq:pasl}) and define the un-annotaetd weights by
\begin{equation}
    \omega_c = \exp(-\alpha \hat{P}_r(c))
\end{equation}
where $\alpha$ is the decay factor. In Table \ref{table: opim_v6_soft_labels} we compare the soft label prior to the configuration used in section \ref{sec:selective_loss}.

% \vspace{0.8cm}
\begin{table}[hbt!]
\centering
\begin{tabular}{ |M{3.6cm}||M{1.3cm}|M{1.3cm}|}
 \hline
 Method & mAP(C) &  mAP(O)\\
 \hline
 P-ASL, \textit{Selective}       & \textbf{86.72}    & 93.57 \\
 P-ASL, \textit{Selective} (soft)                & 86.62   & \textbf{93.59} \\
 \hline
\end{tabular}
\caption{\textbf{OpenImages (V6) results using soft label prior.} We used $\alpha=10$.}
\label{table: opim_v6_soft_labels}
\vspace{-0.2cm}
\end{table}
As the soft label prior provided with lower mAP(C) results, we did not use it in our experiments.

\section{Results on MS-COCO}\label{appendix:b}
\begin{figure}[t!]
\begin{subfigure}[a]{.23\textwidth}
  \centering
  \includegraphics[width=1.0\linewidth]{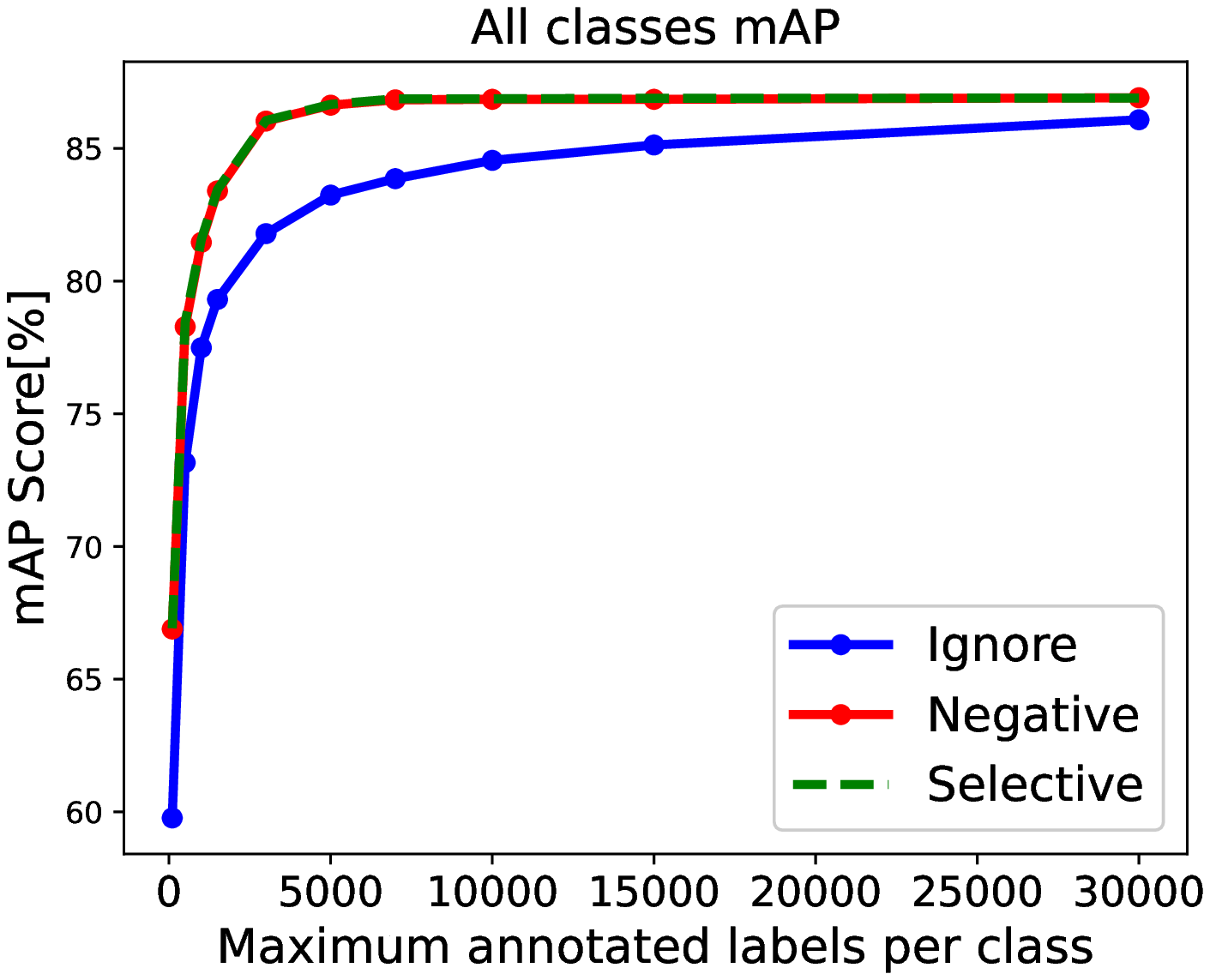}
  \caption{}
\end{subfigure}%
\begin{subfigure}[h]{.23\textwidth }
  \centering
  \includegraphics[width=1.0\linewidth]{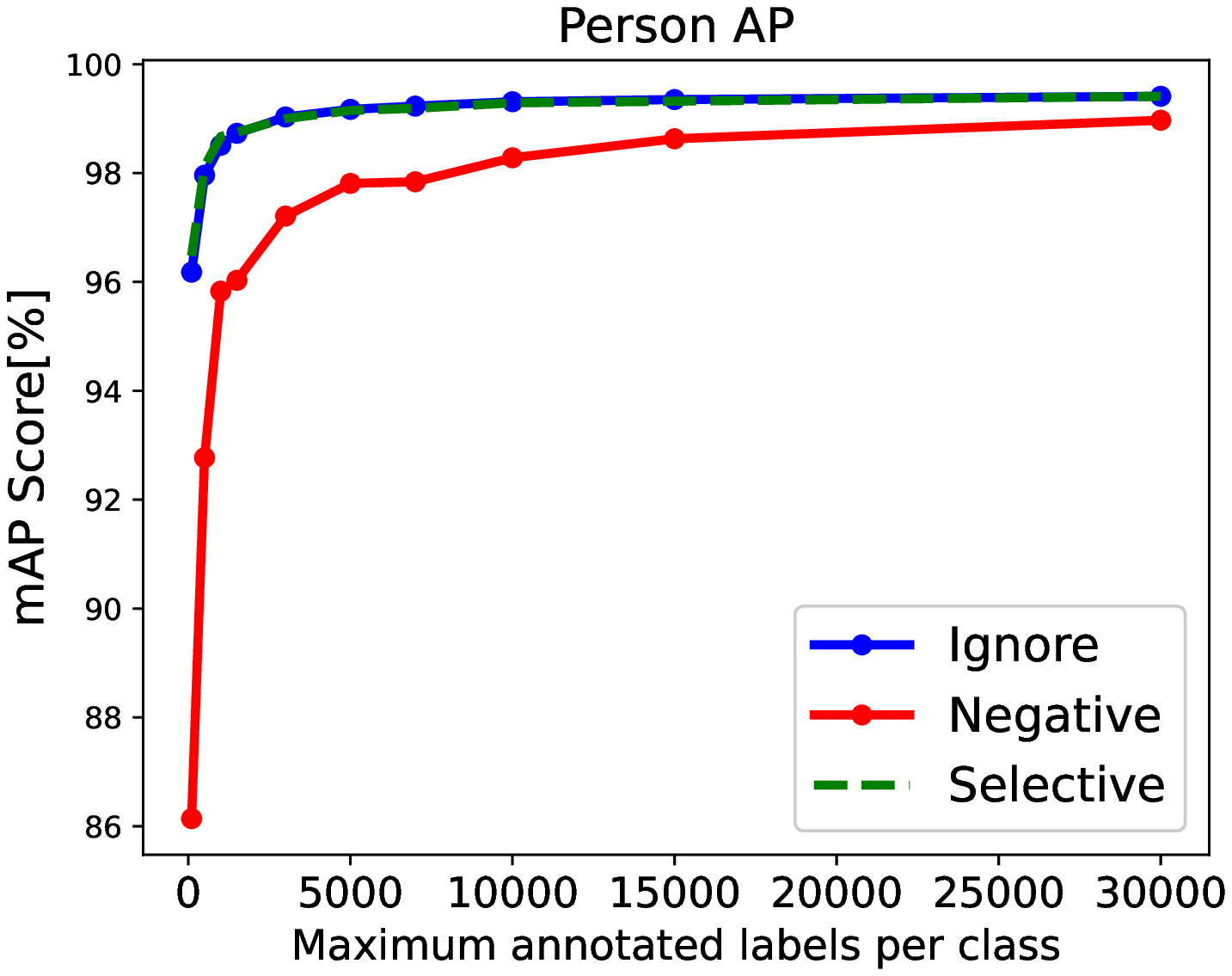}
  \caption{}
\end{subfigure}
\caption{\textbf{Results on MS-COCO (FPC).} .}
\label{fig:coco_selective}
\vspace{-3mm}
\end{figure}

\begin{figure}[t!]
\begin{subfigure}[a]{.23\textwidth}
  \centering
  \includegraphics[width=1.0\linewidth]{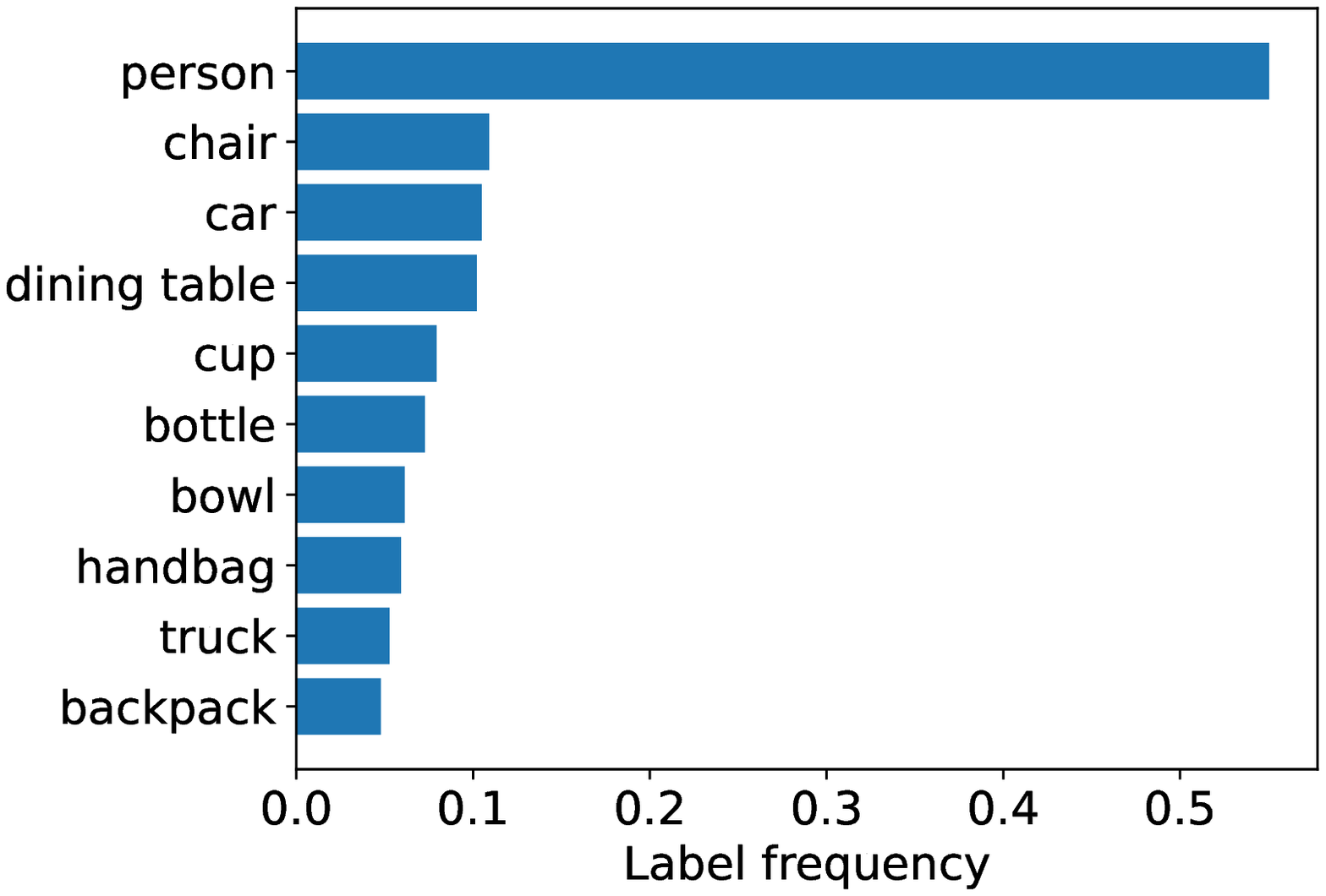}
  \caption{Original top frequencies}
\end{subfigure}%
\begin{subfigure}[h]{.23\textwidth }
  \centering
  \includegraphics[width=1.0\linewidth]{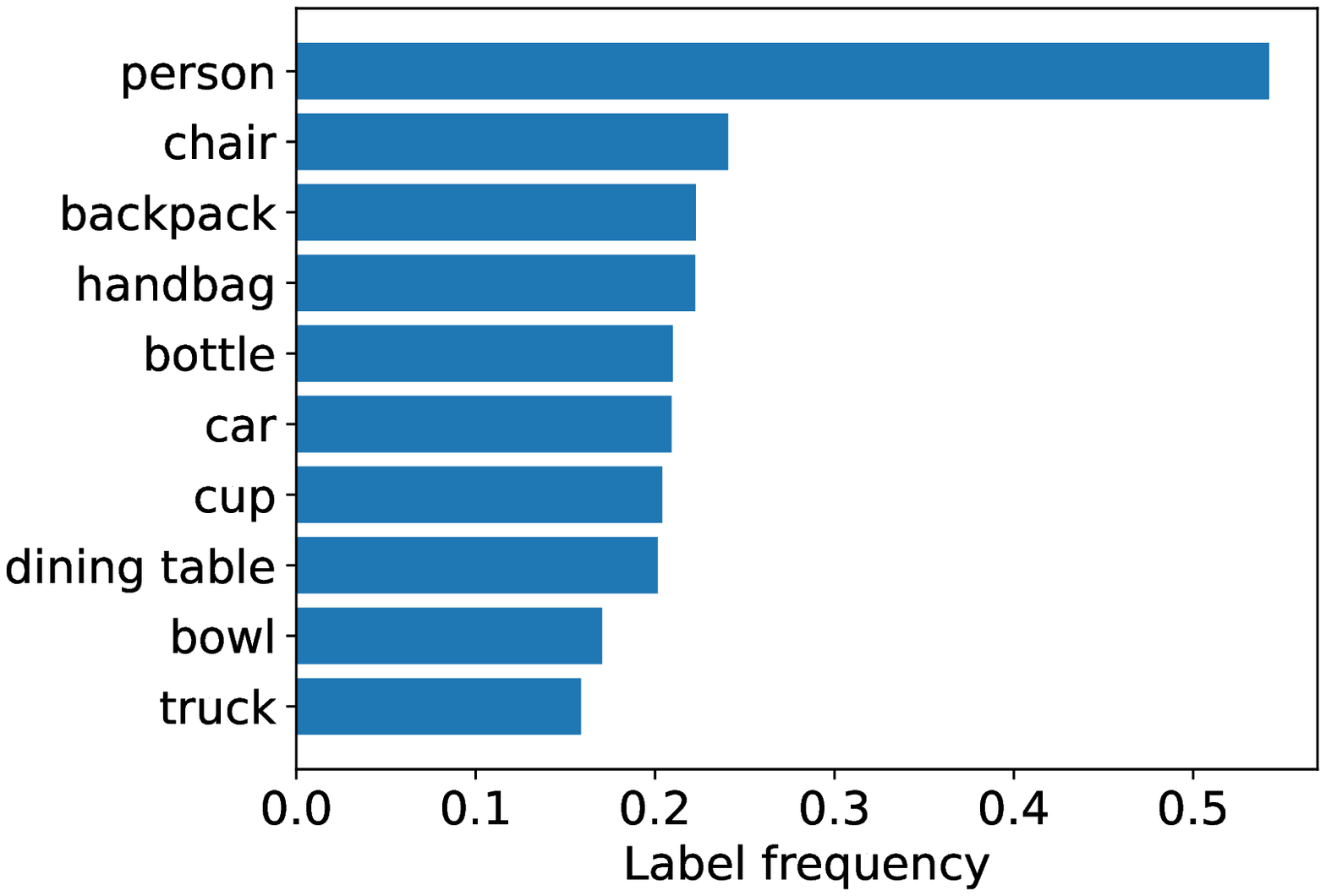}
  \caption{Estimated top frequencies}
\end{subfigure}
\caption{\textbf{Class frequency estimation in MS-COCO.} Top frequent classes measured by (a) original class distribution and (b) estimated class distribution. The estimated top 10 frequent classes are included in the original top classes.}
\label{fig:coco_top_10}
\vspace{-0.3mm}
\end{figure}

In this section, we will present the results obtained on a partially annotated version of MS-COCO, based on the fixed per class (FPC) simulation scheme. Note that in this experiment, the class distribution measured by the number of annotations is no longer meaningful, as all classes have the same number of annotations. The mAP results, as well as the average precision (AP) scores for the class "Person", are presented in Figure \ref{fig:coco_selective}. The \textit{Negative} mode produces higher mAP (computed over all the classes) compared to the \textit{Ignore} mode. However, as the frequent class "Person" is present in most of the images, the \textit{Negative} mode is inferior, especially in the cases of a small number of annotations. Using the \textit{Selective} approach, top results can be achieved for both mAP and the person AP. In Figure \ref{fig:coco_top_10}, we show the top 10 frequent classes obtained using our procedure for estimating the class distribution as described in section \ref{sec: estimate_class_distribution}, and compared them to those obtained using the original class frequencies in MS-COCO. As can be seen, most of the frequent classes measured by the original distribution are also highly ranked by our estimator.

\section{Frequent classes in OpenImages}\label{appendix:c}
\begin{figure*}[t!]
\centering
\begin{subfigure}[a]{.30\textwidth}
  \centering
  \includegraphics[width=1.0\linewidth]{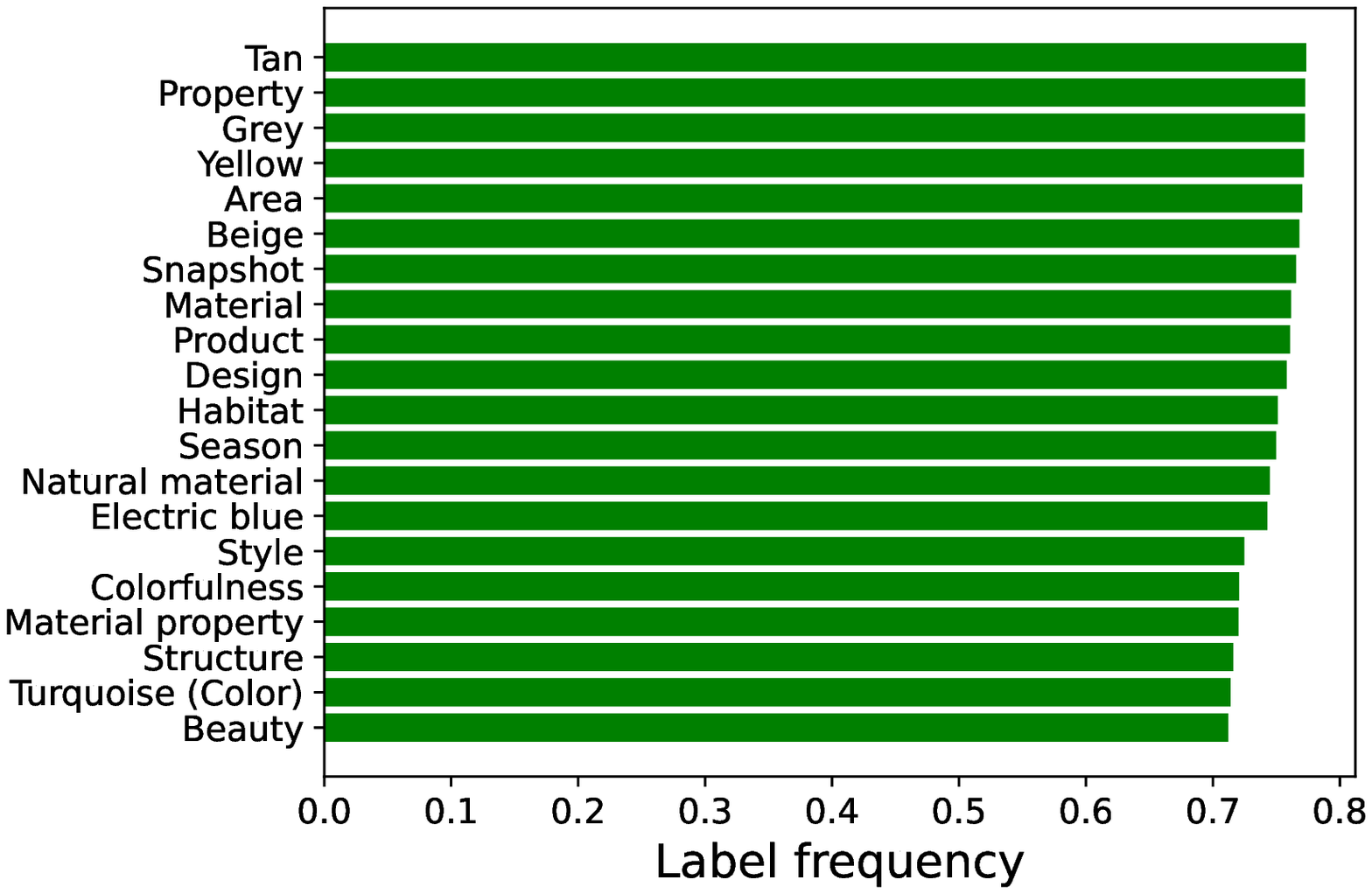}
  \caption{}
\end{subfigure}%
\begin{subfigure}[h]{.30\textwidth }
  \centering
  \includegraphics[width=1.0\linewidth]{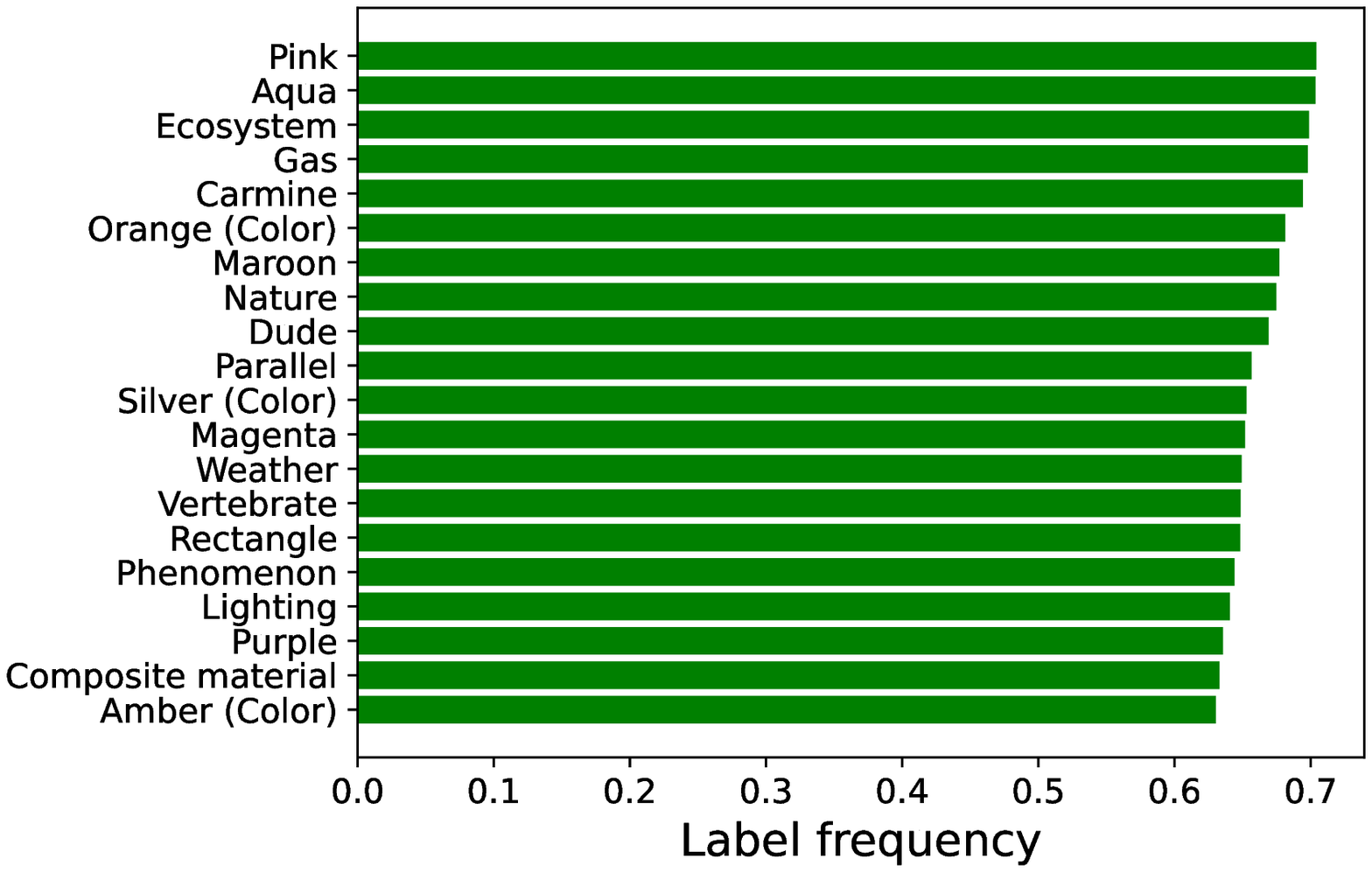}
  \caption{}
\end{subfigure}
\begin{subfigure}[h]{.30\textwidth }
  \centering
  \includegraphics[width=1.0\linewidth]{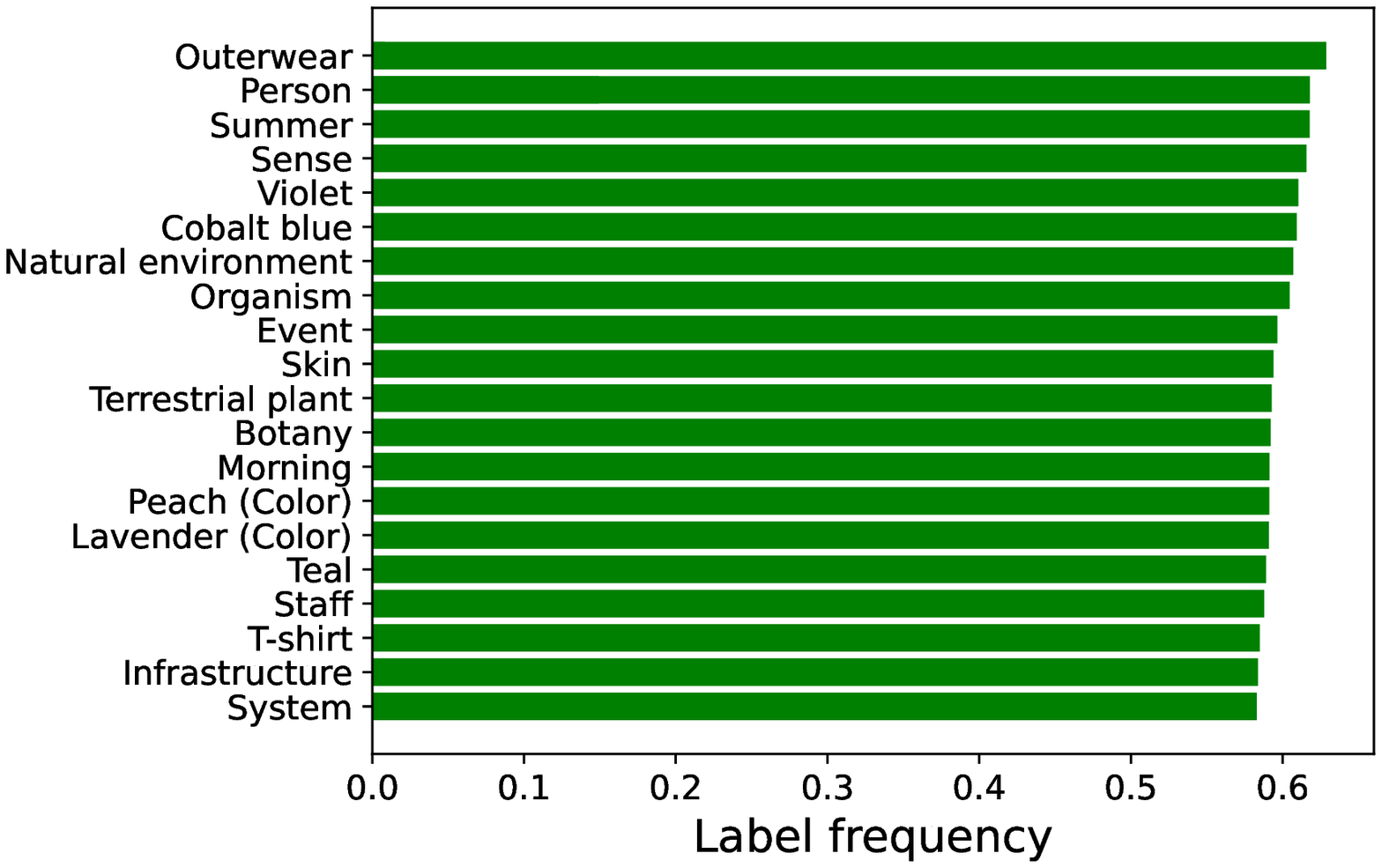}
  \caption{}
\end{subfigure}
\caption{\textbf{Estimating the calss distribution in OpenImages.} Additional top 60 frequent classes as estimated by our approach.}
\label{fig:opim_freq_appendix}
\vspace{-3mm}
\end{figure*}

We add more results of the class distribution estimated by our approach (detailed in \ref{sec: estimate_class_distribution}) for OpenImages dataset. See Figure \ref{fig:opim_freq_appendix}.

\section{Frequent classes in LVIS}\label{appendix:d}
\begin{figure} [t!]
  \vspace{-0.1cm}
  \centering
  \includegraphics[scale=.42]{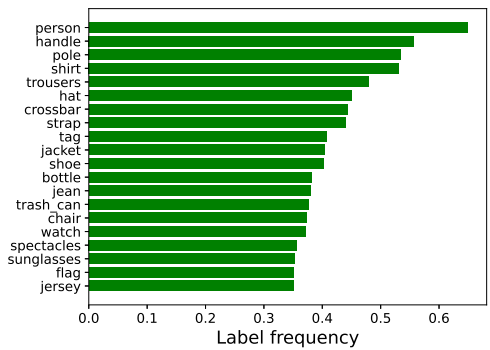}
  \vspace{-0.1cm}
  \caption{\textbf{Estimating the calss distribution in LVIS.} 
%   Common classes may be insufficiently annotated.
  Top 20 frequent classes estimated by the \textit{Ignore} model.}
  \label{fig:estimate_classs_lvis}
  \vspace{-0.1cm}
\end{figure}

In Figure \ref{fig:estimate_classs_lvis} we plot the top frequent classes in LVIS, obtained by our estimator detailed in section \ref{sec: estimate_class_distribution}.
Also in LVIS, it can be seen that the most estimated frequent classes are related to common objects as "Person", "Shirt", "Trousers", "Shoe", etc.

\end{appendices}

\end{document}